\documentclass[acmsmall]{acmart}

\AtBeginDocument{%
  }

\usepackage{amsmath,amssymb,amsfonts}
\usepackage{amsthm}
\usepackage{colortbl}
\usepackage{algpseudocode}
\usepackage{multirow}
\usepackage{algorithm}
\usepackage{graphicx}
\usepackage{textcomp}
\usepackage{booktabs}   
\usepackage{url}
\usepackage{caption}    
\usepackage{enumitem}   
\usepackage{dblfloatfix}
\usepackage{xcolor}
\usepackage{pifont}
\usepackage{adjustbox}

\newcommand{\cmark}{\ding{51}}  
\newcommand{\xmark}{\ding{55}}  

\begin{document}

\title{Learning Wireless Interference Patterns: Decoupled GNN for Throughput Prediction in Heterogeneous Multi-Hop p-CSMA Networks
\\
}

\author{Faezeh Dehghan Tarzjani}
\email{dehghant@usc.edu}
\affiliation{%
  \institution{Dept. of Electrical and Computer Engineering, University of Southern California}
  \city{Los Angeles}
  \state{CA}
  \country{USA}
}

\author{Bhaskar Krishnamachari}
\email{bkrishna@usc.edu}
\affiliation{%
  \institution{Dept. of Electrical and Computer Engineering, University of Southern California}
  \city{Los Angeles}
  \state{CA}
  \country{USA}
}


\begin{abstract}

The p-persistent CSMA protocol is central to random-access MAC analysis, but predicting saturation throughput in heterogeneous multi-hop wireless networks remains a hard problem. Simplified models that assume a single, shared interference domain can underestimate throughput by 48–62\% in sparse topologies. Exact Markov-chain analyses are accurate but scale exponentially in computation time, making them impractical for large networks. These computational barriers motivate structural machine learning approaches like GNNs for scalable throughput prediction in general network topologies. Yet off-the-shelf GNNs struggle here: a standard GCN yields 63.94\% normalized mean absolute error (NMAE) on heterogeneous networks because symmetric normalization conflates a node’s direct interference with higher-order, cascading effects that pertain to how interference propagates over the network graph.

Building on these insights, we propose the Decoupled Graph Convolutional Network (D-GCN), a novel architecture that explicitly separates processing of a node's own transmission probability from neighbor interference effects. D-GCN replaces mean aggregation with learnable attention, yielding interpretable, per-neighbor contribution weights while capturing complex multihop interference patterns.  D-GCN attains 3.3\% NMAE, outperforms strong baselines, remains tractable even when exact analytical methods become computationally infeasible, and enables gradient-based network optimization that achieves within 1\% of theoretical optima.
\end{abstract}


\maketitle

\section{Introduction}
The p-persistent CSMA (p-CSMA) protocol serves as an analytical model for practical random-access MAC protocols \cite{takagi1984throughput} in WLAN and IoT networks, closely replicating the behavior of the basic 802.11 Distributed Coordination Function (DCF) \cite{bianchi2000performance}. In particular, p-CSMA underlies several IEEE 802.11 WLAN design studies that optimize contention windows, frame aggregation, and airtime fairness by treating the channel-access attempt as a Bernoulli trial with probability $p$ per time slot \cite{bononi2004runtime}. 

A natural extension of the classical (homogeneous) p-persistent CSMA protocol is its heterogeneous variant, where each node $i$ contends for the channel with an individual attempt probability $p_i$ \cite{yu2020non}. Allowing per-node probabilities reflects the reality of modern WLANs and IoT deployments, in which devices differ in traffic load, latency requirements, or power constraints \cite{cali2000dynamic}. In heterogeneous networks, nodes experience unequal channel access opportunities based on their local interference environment. Before transmitting, each node must sense whether the channel is idle. When node $i$ transmits, all nodes within its carrier-sense range detect the busy channel and must defer, creating localized contention zones rather than network-wide competition. 

Within this framework, a node can only transmit when all neighbors within its sensing range are silent, either not attempting transmission or themselves blocked by their own neighbors. This leads to each node's throughput depending non-linearly on all transmission probabilities:
\begin{equation}
\Theta_i \approx p_i \cdot f(\{\tilde{p}_j: j \in \mathcal{N}(i)\})
\end{equation}
where $\tilde{p}_j$ represents the effective transmission probability of neighbor $j$, not just its attempt probability $p_j$, but its actual chance to transmit given potential suppression from its own neighbors. Real-world wireless networks exhibit complex multihop interference dependencies. The function $f$ captures both direct interference from immediate neighbors and indirect effects, when a neighbor $j$ is silenced by nodes further away, it cannot interfere with $i$, creating cascading dependencies through the network topology. These multihop interference patterns make $f$ analytically intractable, as it depends recursively on the entire network state.

\begin{figure}[h!]
    \centering
    \includegraphics[width=0.6\linewidth]{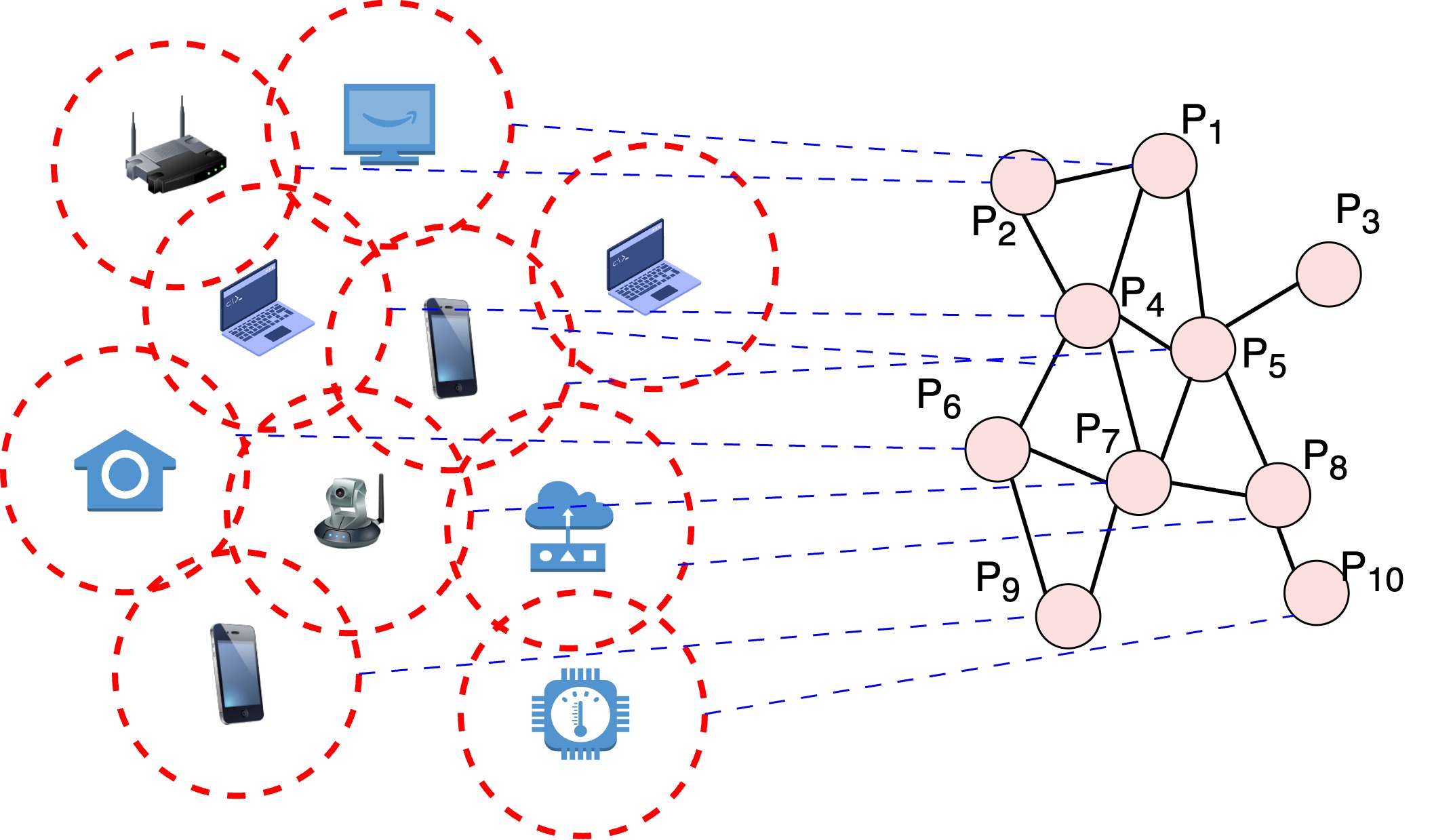}
    \caption{Example of heterogeneous p-persistent CSMA in a 10-node wireless network}
    \Description{Illustration of a 10-node wireless network with mutual interference between nodes modeled by a contention graph.}
    \label{fig:wirelessnetwork}
\end{figure}

This locality of contention is fundamental to practical WLAN/IoT networks. Finite carrier-sense and interference ranges (determined by path loss and shadowing at standard CCA thresholds), physical obstructions, and channelization mean that many node pairs never interact, therefore contention occurs in small, topology dependent zones \cite{abdallah2024improving}. These systems are accurately captured by an undirected conflict graph $G=(V,E)$: vertices denote transmitters, and an edge $(v_i,v_j)\in E$ encodes mutual interference, meaning nodes $i$ and $j$ cannot transmit simultaneously \cite{lu2024graph}. In this representation, $\mathcal{N}(i)$ corresponds to node $i$'s neighbors in $G$ (see Figure~\ref{fig:wirelessnetwork}).

Operators and controllers need models that accurately predict per-node throughput for arbitrary conflict graphs and support optimization of heterogeneous $\{p_i\}$ to meet objectives like proportional fairness, minimum-rate guarantees, or energy/latency trade-offs \cite{yang2024queue}. Traditional renewal theory approaches \cite{gai2011saturation}, which assume complete interference graphs, can underestimate throughput by 48-62\% in sparse topologies. Recently, an exact Markov chain method to compute the throughputs for arbitrary topologies correctly has been introduced~\cite{11133870}. However, this approach suffers from state-space explosion as the underlying chain has $T^n$ states (where $T$ is the transmission duration and $n$ the network size). For instance, a modest network with $n=10$ nodes and $T=5$ yields $10^7$ states, rendering exact analysis computationally intractable for optimization tasks that require repeated throughput evaluations \cite{afifi2024machine, sun2025comprehensive}.


These computational barriers and the inherent graph structure of interference patterns naturally motivate a learning-based approach. Graph Neural Networks offer compelling advantages for this domain: they operate directly on graph-structured conflict topologies while learning complex non-linear relationships between transmission probabilities and throughputs. Through iterative message passing, GNNs capture multihop interference dependencies, and critically, they provide differentiable throughput predictions that enable gradient-based optimization of network parameters. 

Despite this natural alignment, existing applications of machine learning to CSMA protocol optimization have primarily focused on simple scenarios or relied on architectures that fail to exploit the structural properties of interference graphs. Most prior work either assumes simplified conflict models or employs generic neural architectures without incorporating domain-specific inductive biases crucial for wireless network modeling.

This paper addresses these limitations by introducing a new Graph Neural Network architecture tailored to heterogeneous p-CSMA networks that we refer to as Decoupled Graph Convolutional Network (D-GCN). Our aim is to deliver accurate per-node throughput prediction and enable efficient optimization on arbitrary conflict-graph topologies across heterogeneous access probabilities $\{p_i\}$ and packet durations $T$.

Our D-GCN approach introduces three key innovations that distinguish it from standard GNN architectures:

First, D-GCN decouples self-transmission from neighbor interference processing, mirroring the multiplicative structure of $p$-CSMA throughput seen in simpler topologies such as complete graphs: $\Theta_i \approx p_i \cdot f(\{\tilde{p}_j: j \in \mathcal{N}(i)\})$. Standard GNNs mix these fundamentally different signals before projection, obscuring the distinction between a node's transmission capability and the suppression it experiences. 

Second, D-GCN eliminates degree normalization prevalent in GCN and GraphSAGE, which incorrectly dilutes cumulative interference by averaging neighbor contributions. In wireless networks, interference is additive---more active neighbors mean more contention, not averaged impact. Our architecture preserves this additive nature through unnormalized summation with learnable attention weights. 

Third, D-GCN's multi-layer architecture captures k-hop interference cascades, where each layer extends the interference horizon by one hop. This provides a computationally tractable alternative to the exponentially complex analytical methods required for modeling these spatial-temporal dependencies.

We demonstrate that our approach achieves high prediction accuracy (normalized mean absolute error below 8\%) across diverse network configurations and scales gracefully with network size and transmission duration. Our learned model serves as an effective surrogate for exact analytical methods in optimization applications, achieving utility values within 1\% of theoretical optima. The methodology provides a general framework for applying Graph Neural Networks to wireless protocol optimization, demonstrating how domain-specific architectural modifications can significantly improve performance on structured prediction tasks.

The remainder of this paper is organized as follows: Section 2 reviews related work in CSMA analysis, network utility maximization, and Graph Neural Networks for wireless communications. Section 3 formally defines the problem and establishes the computational challenges motivating our approach. Section 4 presents our proposed GNN architectures and compares them with existing approaches. Section 5 describes our dataset generation methodology and evaluation metrics. Section 6 provides comprehensive experimental results demonstrating the effectiveness of our approach across multiple dimensions. Finally, Section 7 concludes with discussion of implications and future research directions.

\section{Related Work}

The challenge of analyzing and optimizing throughput in CSMA networks has attracted significant research attention from both analytical and learning-based perspectives.
Bianchi's seminal work \cite{bianchi2000performance} introduced a two-dimensional Markov chain model for IEEE 802.11 DCF, accurately predicting saturation throughput for homogeneous nodes in a single collision domain. For IEEE 802.15.4 MAC, Ling et al. \cite{ling2008renewal} developed a renewal-theoretic model (slotted non-persistent CSMA with BEB) and derive normalized saturation throughput and frame service time for saturated nodes;  Gai, Ganesan, and Krishnamachari \cite{gai2011saturation} compute the exact per node throughput for single domain and  characterize the saturation throughput region of slotted p-persistent CSMA, providing a closed-form Pareto boundary. 
However, these early analytical approaches presumed simplified interference models assuming single collision domains where all nodes interfere with each other, failing to capture the complexity of real-world scenarios.

Jiang and Walrand \cite{jiang2009distributed} developed a distributed CSMA algorithm that achieves optimal throughput using Gibbs sampling, establishing the connection between CSMA scheduling and maximum weight independent set problems. Their approach demonstrates that CSMA can implicitly solve NP-hard optimization problems, but the exact computation still requires exponential complexity $O(T^n 2^n)$ for n nodes with transmission duration T.
Arthi and Mehta \cite{arthi2024saturation} analyze saturation throughput for a hybrid access MAC in IEEE 802.11ax (Wi-Fi 6), where scheduled OFDMA access coexists with random access; they develop an analytical model and validate it with numerical results. Their study targets single-cell WLAN operation and standard-specific features rather than arbitrary conflict graphs or heterogeneous $p_i$.
Recent work has addressed this limitation by developing exact computational approaches based on novel Markov chain formulations that can handle arbitrary conflict graph topologies \cite{11133870}, these methods face fundamental scalability challenges due to exponential state space growth with network size.
Tarzjani and Krishnamachari \cite{11133870} revealed a critical limitation of traditional renewal theory approaches, demonstrating a 48-62\% throughput underestimation in sparse conflict graphs, a devastating finding for practical IoT deployments. Their exact Markov chain formulation provides accurate results but still faces a $O(T^n)$ state space explosion.
\par
The network utility maximization (NUM) field has successfully transitioned from theoretical frameworks to practical implementations that handle real-world wireless complexity, and the integration of machine learning has fundamentally transformed NUM's ability to handle real-world complexity \cite{cao2022online}. 
 Learning-based approaches now address unknown utilities \cite{ji2023network}.

\par
Graph Neural Networks have emerged as the dominant paradigm for wireless network optimization with optimal performance while providing 1000x speedup over traditional iterative methods \cite{moorthy2024survey}. The research landscape shows Graph Convolutional Networks (GCN), Graph Attention Networks (GAT), and GraphSAGE as the primary architectures, each excelling in different contexts. GCN offers sub-millisecond inference, making it ideal for real-time tasks like power control \cite{wang2023engnn}.
GAT improves precision by 1.25–3.04\%, excelling in complex  management scenarios \cite{sun2025jumping}.
GraphSAGE scales efficiently, sustaining 98\% optimal performance even in networks 5× larger than training \cite{shen2022graph}.
The architecture selection follows clear patterns based on application requirements. MAC protocol optimization using GNNs represents the most underdeveloped area despite substantial theoretical potential. While GNNs have shown promise in some wireless applications, their use for CSMA protocol optimization remains severely limited.
CSMA protocol optimization using GNNs remains largely unexplored, 
Moon et al.\cite{moon2021neuro} proposed 
Neuro-DCF, which combines MARL with GNN to learn adaptive CSMA policies, 
demonstrating significant delay reduction while maintaining throughput 
optimality. However, Neuro-DCF employs GNN as a feature extractor within 
a complex MARL framework rather than as the primary optimization mechanism.

To our knowledge, no prior work has investigated GNN architectures for heterogeneous p-CSMA throughput prediction and optimization. While the baseline architectures (GCN, GraphSAGE, GIN, GINE) are well-established in the GNN literature, we are the first to implement and evaluate them for this wireless networking problem, along with our newly proposed D-GCN architecture.
\section{Problem Definition}

Consider a wireless network that employs the \emph{heterogeneous $p$-persistent CSMA} (p-CSMA) medium--access protocol. Let $V=\{1,\dots,n\}$ index saturated transmitters. Time is divided into slots of unit length (we set $\sigma\!=\!1$ without loss of generality.)

\begin{figure}[h!]
    \centering
    \includegraphics[width=0.4\linewidth]{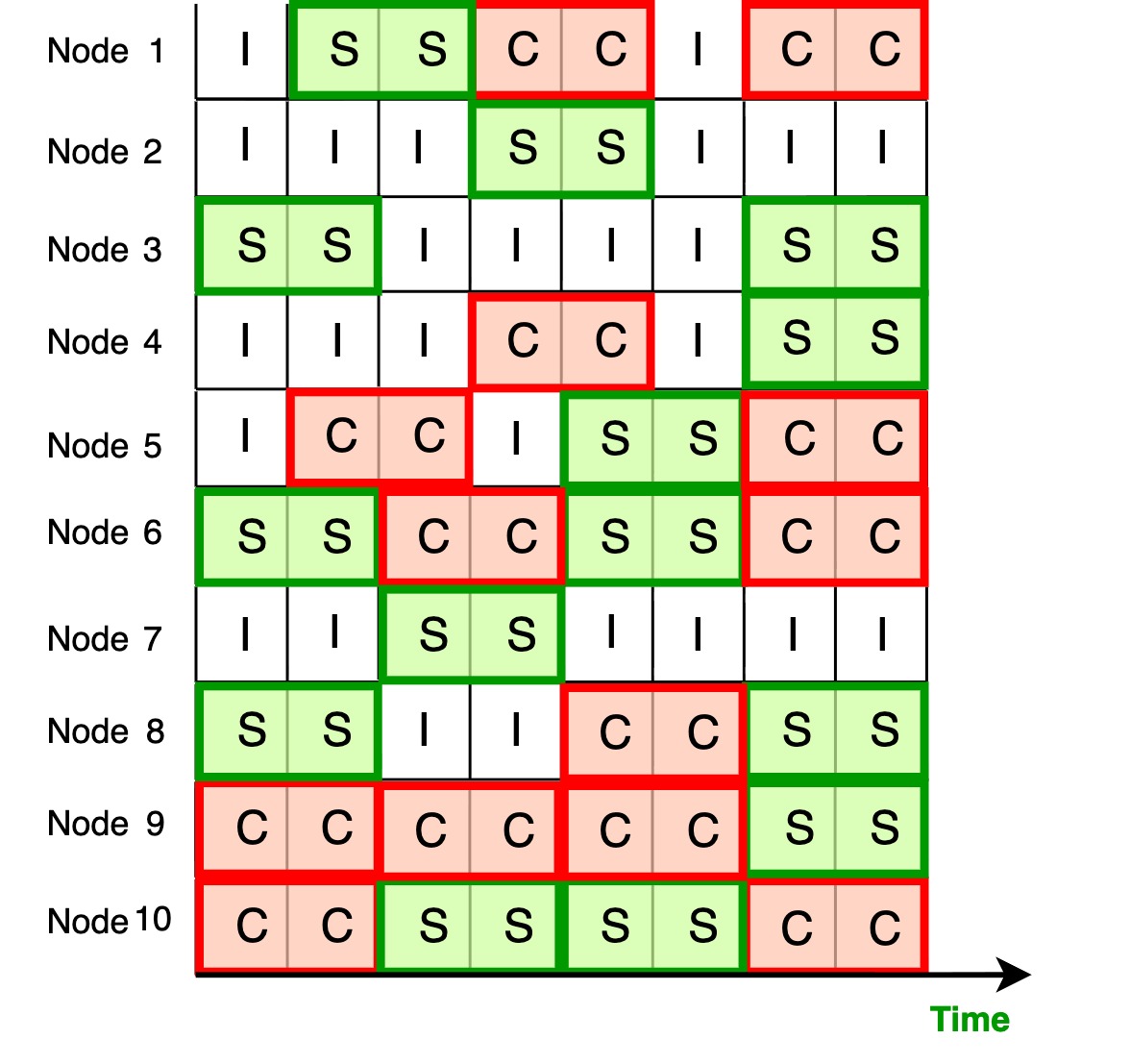}
    \caption{Example time slot assignment for Figure~\ref{fig:wirelessnetwork} p-CSMA network with transmission duration T=2. Each row represents a node's channel access pattern over 8 time slots. Green boxes (S) indicate successful transmission, red boxes (C) denote conflicts where transmission attempts fail due to neighbor interference, and white boxes (I) represent idle slots. }
    \label{fig:timeslot}
\end{figure}

If a node senses that the channel is idle at the beginning of a time slot, it attempts transmission with its own Bernoulli probability $p_i \in [0,1]$; otherwise, it defers for one slot. Once a node begins transmission, it occupies the channel for $T$ consecutive time slots. If two neighboring nodes simultaneously sense the channel as idle and initiate transmission, a collision occurs for $T$ consecutive time slots. Figure~\ref{fig:timeslot} illustrates a representative scenario for the wireless network and conflict graph shown in Figure~\ref{fig:wirelessnetwork}.

The \emph{saturation throughput} of node $i$ is the long-run fraction of time slots it holds the channel:
\begin{equation}
S_i(t) \;:=\; \mathbf{1}\{\text{$i$ begins a \emph{collision-free} transmission at slot $t$}\}.
\end{equation}
\begin{equation}
\Theta_i \;=\; T \cdot \lim_{L\to\infty}\frac{1}{L}\sum_{t=0}^{L-1} S_i(t).
\label{eq:theta-def}
\end{equation}

The computational challenge of heterogeneous p-CSMA networks manifests at two levels:

\noindent\textbf{State space and scalability.}
We represent the system at time slot $t$ by the residual-timer vector
$s(t)=(a_1(t),\dots,a_n(t))\in\{0,\dots,T-1\}^{n}$, where $a_i(t)=0$ means node $i$
is idle/eligible and $a_i(t)>0$ are the remaining time slots that it must stay busy.
Only idle nodes attempt; a node succeeds if and only if it attempts while none of its
graph neighbors do, so winners form an independent set of the undirected
conflict graph $G=(V,E)$. Timers update synchronously, winners reset to $T-1$,
while the others decrement to $\max\{a_i(t)-1,0\}$.
This yields a finite, time-homogeneous Markov chain on
$\mathcal{S}=\{0,\dots,T-1\}^{n}$ with size $|\mathcal{S}|=T^{\,n}$.

From any state $s$, one-step transitions require summing over the $2^{|I(s)|}$
attempt patterns of the idle set $I(s)=\{i:a_i=0\}$ (worst case $2^{|I(s)|}\le 2^{n}$).
Constructing the kernel $P$ and solving $P^\top\pi=\pi$ therefore scales on the
order of $O(T^{\,n}2^{n})$ time and $O(T^{\,n})$ memory, which explodes rapidly
with $T$ and $n$ (e.g., $T{=}5,n{=}12\Rightarrow |\mathcal{S}|=5^{12}\approx2.44\times10^8$;
$T{=}7,n{=}12\Rightarrow 7^{12}\approx1.38\times10^{10}$), making exact analysis
impractical beyond small networks or packet durations.

\noindent\textbf{Optimization complexity.}
The state-space explosion described above means that computing $\Theta_i(\mathbf{p})$ for any given probability vector $\mathbf{p}$ is already computationally expensive. This evaluation challenge becomes particularly problematic when embedded within an optimization framework, where throughput must be evaluated repeatedly for different probability configurations. Specifically, throughput-based network utility optimization in heterogeneous p-CSMA networks involves finding optimal transmission probabilities to maximize network performance while respecting interference constraints. The saturation throughput-based network utility optimization problem can be formulated as follows:
\begin{align*}
\text{maximize} \quad & \sum_{i=1}^{n} \alpha_i U(\Theta_i(p_1, p_2, \dots, p_n)) \\
\text{subject to} \quad & p_i \in [0,1], \quad \forall i \in \{1,2,\dots,n\} \\
& g_k(p_1, p_2, \dots, p_n) \leq 0, \quad k = 1, \ldots, K \\
& \Theta_i(p_1, p_2, \dots, p_n) \geq \Theta_i^{\min}
\end{align*}
Here, $\Theta_i(\cdot)$ represents the saturation throughput of node $i$, which is a complex non-linear function of all transmission probabilities $\mathbf{p} = \{p_1, p_2, \dots, p_n\}$, the network topology and $T$ transmission duration. The optimization variable $p_i$ denotes the transmission probability of node $i$ in the p-CSMA protocol, bounded within $[0,1]$ to ensure valid probabilities. The function $U(\cdot)$ is a utility function (e.g., log utility for proportional fairness), and $\alpha_i \geq 0$ are weights that allow for different priority assignments to nodes. The constraints $g_k(\cdot)$ encompass fairness, stability, interference limits, minimum throughput $\Theta_i^{\min}$, capacity bounds, and QoS requirements.

Selecting the access probabilities $\mathbf{p}$ that maximize a utility of the resulting throughputs reduces to a \emph{weighted} \textsc{Maximum Independent Set} on the conflict graph, an NP-hard problem.  
These dual hurdles---state-space explosion (worsened by larger $T$ and network scale) and combinatorial optimization---explain why exact saturation-throughput evaluation and tuning become computationally intractable once the network departs from the single-collision-domain idealization.

\section{Proposed Methodology}
Our approach is motivated by three key observations. 
First, exact methods, while accurate, become computationally intractable beyond modest network sizes due to complexity 
$\mathcal{O}\!\left(T^n 2^n\right)$. 
Second, traditional analytical approximations fail catastrophically on non-complete conflict graphs, making them unsuitable for real-world deployments. 
Finally, the inherent graph structure of wireless interference patterns naturally aligns with GNN's message passing paradigm, enabling efficient learning of complex spatial dependencies.

\begin{figure}[h]
    \centering
    \includegraphics[width=1\textwidth]{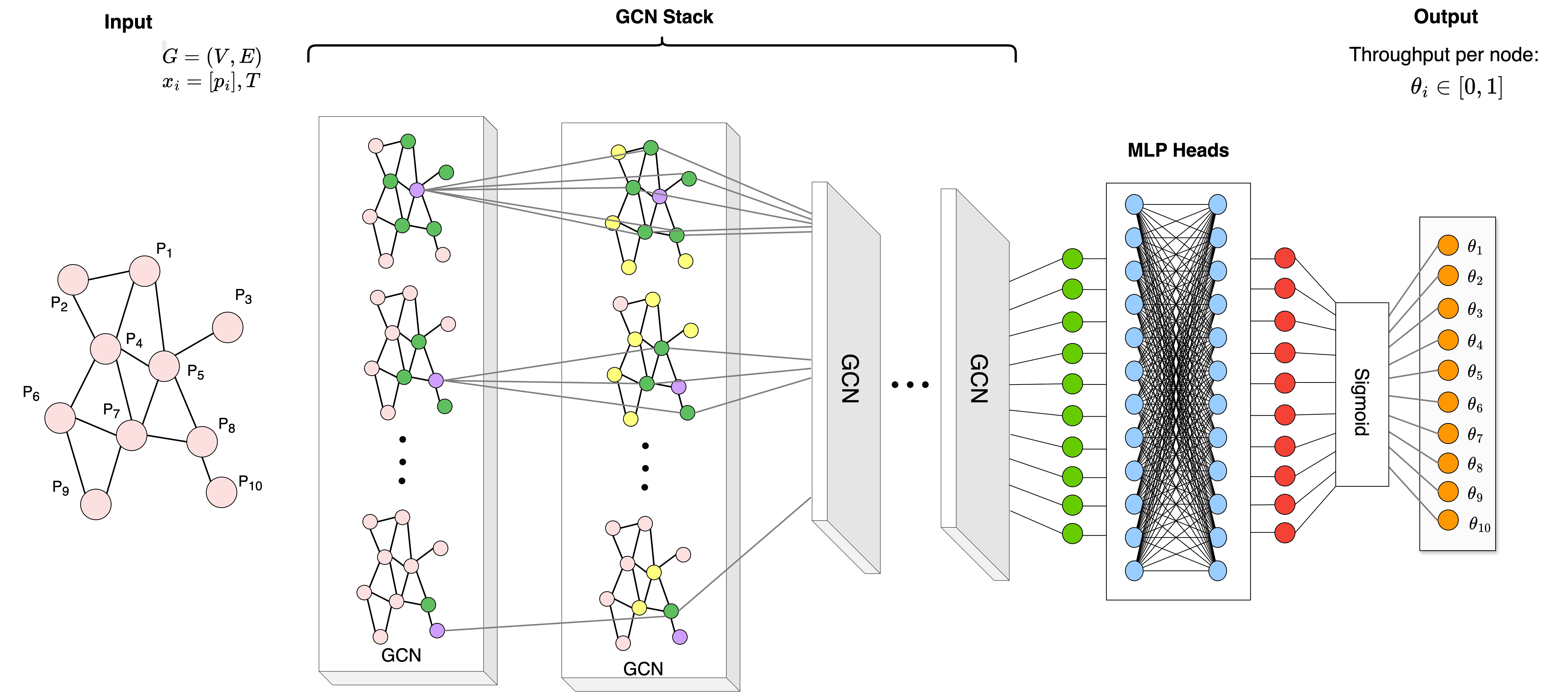}
    \caption{Proposed Graph Neural Network architecture for heterogeneous p-CSMA throughput}
    \label{fig:GCNarch}
\end{figure}

As illustrated in Figure~\ref{fig:GCNarch}, In our proposed GNN architecture for this problem, information propagates through the network via iterative message passing layers. A generic message passing layer at depth $\ell$ performs the following computation:

\begin{equation}
h_v^{(\ell+1)} = \mathrm{UPD}^{(\ell)}\!\Big(
    h_v^{(\ell)},\;
    \mathrm{AGG}^{(\ell)}\{\,\mathrm{MSG}^{(\ell)}(h_v^{(\ell)},h_u^{(\ell)},e_{uv})\Big).
\label{eq:mpnn-template}
\end{equation}

Here, $h_v^{(\ell)}$ represents the hidden state of node $v$ at layer $\ell$ 
(initialized as $h_v^{(0)} = x_v$), $\mathcal{N}(v)$ denotes the neighbors of node $v$ 
in the conflict graph, and $e_{uv}$ represents edge features.
The \textsc{Message} function computes pairwise interactions, 
\textsc{Aggregate} combines messages from all neighbors, 
and \textsc{Update} integrates this aggregated information with the node's current state.

After $L$ message passing layers that progressively capture multihop interference patterns, 
we apply a Multi-Layer Perceptron (MLP) head to each node's final representation:
\[
\hat{\Theta}_v = \sigma\!\Big(\mathrm{MLP}(h_v^{(L)})\Big).
\]

The MLP head consists of two fully-connected layers and ReLU activations between layers. 
The final sigmoid activation $\sigma(\cdot)$ ensures the predicted throughput 
$\hat{\Theta}_v \in [0,1]$, respecting the physical constraint that throughput cannot exceed channel capacity. 

This architecture serves dual purposes:  
First, it transforms the graph-aware embeddings into task-specific throughput predictions;  
Second, it increases model capacity without requiring additional GNN layers, 
which would risk over-smoothing due to excessive neighborhood aggregation.

\subsection{Baseline GNN Architectures for Heterogeneous Interference Modeling}
To understand the architectural requirements for modeling wireless interference, we adapt and compare five GNN variants that differ in their aggregation schemes and neighbor information processing:
GCN (Graph Convolutional Network)~\cite{kipf2016semi} employs symmetric normalization and mixed self/neighbor transformations. 
GraphSAGE~\cite{hamilton2017inductive} introduces separate self/neighbor weights with mean aggregation. 
GIN (Graph Isomorphism Network)~\cite{xu2018powerful} uses summation aggregation with learnable self-weighting. 
GINE (GIN with Edge features)~\cite{hu2019strategies} extends GIN with edge-aware message passing. 
Finally, D-GCN (Decoupled GCN), our proposed architecture, incorporates attention-weighted neighbor suppression.

Each architecture instantiates the general message passing framework from Equation~\eqref{eq:mpnn-template} differently, leading to distinct inductive biases that we now examine in detail.

\subsubsection{GCN (Kipf \& Welling)}
\label{sec:arch-gcn}
The Graph Convolutional Network~\cite{kipf2016semi} applies symmetric degree normalization:
\begin{align}
\tilde{A} = A + I, \qquad
\tilde{D}_{vv} = \sum_{u} \tilde{A}_{vu}, \\
h^{(\ell+1)} = \sigma\!\left(\tilde{D}^{-1/2} \tilde{A} \tilde{D}^{-1/2}
                     h^{(\ell)} W^{(\ell)}\right),
\end{align}

GCN \emph{averages} neighbor signals (scaled by degree) and applies a single
linear transform; self and neighbor information are mixed before projection.
In contention graphs, this tends to \emph{over-smooth} high-contrast local load
signals, contributing to underfitting.

\subsubsection{GraphSAGE (Hamilton et al.)}
\label{sec:arch-sage}
The Graph Sample and Aggregate (GraphSAGE) architecture~\cite{hamilton2017inductive} introduces explicit separation between self and neighbor transformations:
\begin{equation}
\scalebox{0.8}{$
h_v^{(\ell+1)} = \sigma\Bigl(
    W_{\mathrm{self}}^{(\ell)} h_v^{(\ell)} + W_{\mathrm{neigh}}^{(\ell)} \cdot \mathrm{mean}_{u \in \mathcal{N}(v)} h_u^{(\ell)}
\Bigr),
$}
\end{equation}
where $W_{\mathrm{self}}$ and $W_{\mathrm{neigh}}$ are separate weight matrices for self and neighbor transformations. While this decoupling improves upon GCN's mixed transformations, the mean aggregation normalizes neighbor contributions by degree, which our experiments show is detrimental for heterogeneous interference graphs.

\subsubsection{GIN (Xu et al.)}
\label{sec:arch-gin}

The Graph Isomorphism Network (GIN)~\cite{xu2018powerful} matches the discriminative power of the 1-Weisfeiler-Lehman graph isomorphism test and outperforms GCN and GraphSAGE on several benchmarks. The Graph Isomorphism Network replaces degree-normalized averaging with degree-agnostic summation
followed by an MLP, and introduces a learnable self-weight $\epsilon^{(\ell)}$:
\begin{equation}
h_v^{(\ell+1)} =
\mathrm{MLP}^{(\ell)}\!\Big(
  (1+\epsilon^{(\ell)})\,h_v^{(\ell)} \;+\;
  \sum_{u\in\mathcal{N}(v)} h_u^{(\ell)}
\Big).
\end{equation}
Summation treats each neighbor equally while preserving multiset counts; the MLP
can learn highly non-linear functions of the aggregated local load
($\sum p_u$, etc.), which is critical for approximating collision probability.

\subsubsection{GINE (Hu et al.)}
\label{sec:arch-gine}

The Graph Isomorphism Network with Edge features~\cite{hu2019strategies} incorporates edge attributes additively inside a ReLU before summation:

{
\begin{equation}
\scalebox{0.8}{$
\begin{aligned}
h_v^{(\ell+1)} &=
  \mathrm{MLP}^{(\ell)}\!\Bigl(
      (1+\epsilon^{(\ell)})\,h_v^{(\ell)}
      \;+\;
      \sum_{u\in\mathcal{N}(v)}
          \mathrm{ReLU}\!\bigl(
              h_u^{(\ell)} + \phi(e_{uv})
          \bigr)
  \Bigr).
  \\[4pt]
  \phi(e_{uv}) &= W_e\,e_{uv} + b_e,
\end{aligned}
$}
\end{equation}
}

When $e_{uv}$ encodes link strength or interference severity, GINE can learn
to modulate neighbor impact. With binary edges (our data), this reduces to a
learnable shift/gating that nonetheless improves stability versus vanilla GIN.

While evaluating various GNN architectures, we identified critical limitations that fundamentally misalign with the unique characteristics of wireless interference networks. 
Standard GNN architectures fail to capture the unique dynamics of wireless interference networks. When architectures combine self and neighbor information before projection, they obscure the fundamental distinction between a node's transmission capability and the suppression it experiences, making it harder to learn how each neighbor's transmission probability, position in the topology, and their own neighborhood structure affects interference.
Mean aggregation schemes incorrectly dilute cumulative interference effects, while mixed transformations lack the interpretability needed for protocol optimization.

Also, lack of interpretability is particularly problematic for protocol optimization, where understanding which neighbors cause the most contention is essential for parameter tuning.

\subsection{Decoupled Graph Convolutional Network (D-GCN)}
 
\label{sec:arch-enhancedgcn}

To address these limitations, we propose the Decoupled Graph Convolutional Network (D-GCN), 
which incorporates four key architectural innovations: (i) explicit separation of self-transmission 
and neighbor interference processing channels, (ii) elimination of degree normalization and mean 
aggregation that incorrectly dilute cumulative interference effects, (iii) learnable attention 
weights to capture heterogeneous neighbor impacts, and (iv) unnormalized summation that preserves 
the additive nature of wireless interference—where more active neighbors create more contention, 
not averaged impact.

\begin{equation}
\begin{aligned}
z_u^{(\ell)} &= h_u^{(\ell)} W_{\text{nbr}}^{(\ell)} \quad \text{(neighbor transformation)}\\
\alpha_{uv}^{(\ell)} &= \sigma\big(\mathbf{a}^{(\ell)\top} z_u^{(\ell)}\big) \quad \text{(learned importance)}\\
h_v^{(\ell+1)} &= \sigma\Bigg(\underbrace{h_v^{(\ell)} W_{\text{self}}^{(\ell)}}_{\text{self channel}} + \underbrace{\sum_{u \in \mathcal{N}(v)} \alpha_{uv}^{(\ell)} \cdot \text{ReLU}(z_u^{(\ell)})}_{\text{neighbor suppression}} + b^{(\ell)}\Bigg)
\end{aligned}
\end{equation}

The decoupled weight matrices $W_{\text{self}}$ and $W_{\text{nbr}}$ are learnable linear 
transformations that extract different feature representations depending on their role in the 
message-passing process. At each layer $\ell$, $W_{\text{self}}^{(\ell)}$ processes a node's 
own state to learn how its current embedding translates to channel access capability, while 
$W_{\text{nbr}}^{(\ell)}$ processes neighbor states to learn their interference contribution. 
Critically, through $L$ stacked layers, this architecture captures \emph{multi-hop interference 
cascades}: Layer 1 processes direct (1-hop) neighbors, Layer 2 incorporates information from 
2-hop neighbors (neighbors of neighbors), and Layer $L$ can theoretically capture interference 
dependencies up to $L$ hops away. This architectural design directly mirrors the fundamental throughput relationship in saturated CSMA, where $\Theta_i \approx p_i \cdot f(\{\tilde{p}_j: j \in \mathcal{N}(i)\})$, the node's throughput is its attempt probability modulated by a suppression factor from interfering neighbors.

The learned attention weights $\alpha_{uv} = \sigma(\mathbf{a}^T z_u)$ serve as a soft, data-driven interference mask that captures the heterogeneous impact of different neighbors. Unlike uniform aggregation schemes, this mechanism allows the model to learn that some neighbors may cause stronger interference than others based on their transmission patterns. The ReLU activation applied before aggregation ($\sum_{u \in \mathcal{N}(v)} \alpha_{uv} \cdot \text{ReLU}(z_u)$) provides non-linearity essential for learning complex interference patterns, without it, multiple linear layers would collapse to a single linear transformation. We adopt 
this design from GINE \cite{hu2019strategies}, which applies ReLU before aggregation to enable 
element-wise non-linear transformations of neighbor features. In wireless networks, interference 
relationships are inherently non-linear due to collision dynamics and temporal dependencies, 
making non-linear activations crucial for accurate throughput prediction.

\begin{figure}[h!]
    \centering
    \includegraphics[width=0.45\linewidth]{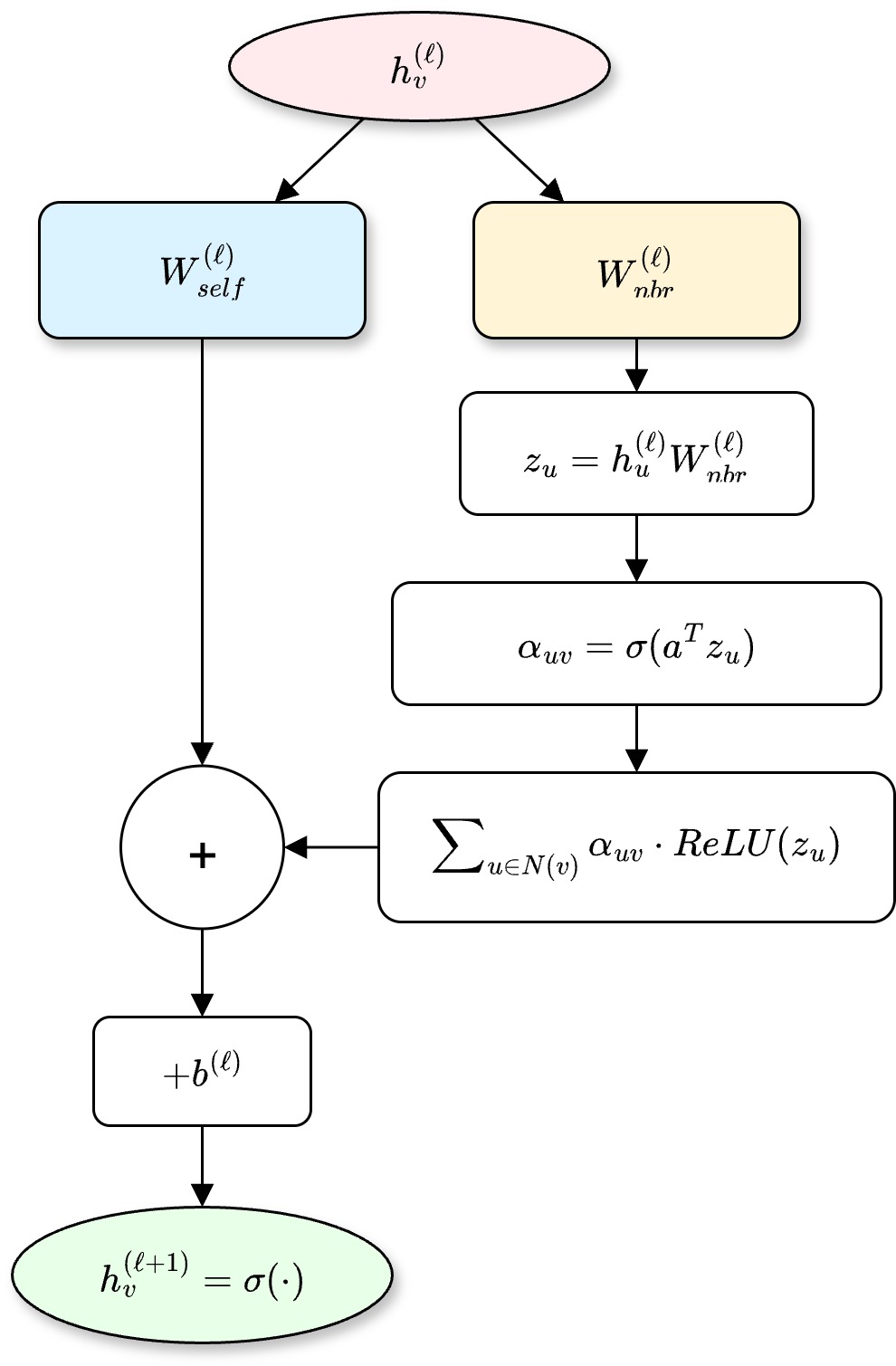}
    \caption{Architecture of the Decoupled Graph Convolutional Network (D-GCN) for p-CSMA throughput prediction.}
    \label{fig:Architecture}
\end{figure}

Figure~\ref{fig:Architecture} visualizes this computation flow. The node's embedding $h_v^{(\ell)}$ 
follows two parallel paths: the self-channel (left) directly transforms the node's features via 
$W_{\text{self}}$, while the neighbor path (right) processes each neighbor through transformation 
($W_{\text{nbr}}$), attention weighting ($\alpha_{uv}$), and ReLU non-linearity before aggregation. 
These pathways are summed with bias $b^{(\ell)}$ and activated to produce the next-layer embedding 
$h_v^{(\ell+1)}$.

\subsection{Summary of Architectural Differences}
\label{sec:arch-summary}
Table~\ref{tab:arch-comparison} summarizes the key architectural distinctions among the evaluated models. These design choices influence the ability of each architecture to capture the multiplicative nature of wireless interference, ensure stable training in heterogeneous networks, and exhibit heterophily robustness (handling dissimilar connected nodes) as well as over-smoothing resistance (preserving distinct node representations in deeper layers).

\begin{table}[h]
\centering
\caption{Architectural feature comparison highlighting distinctions between baseline GNNs and the proposed D-GCN}
\label{tab:arch-comparison}
\begin{adjustbox}{max width=\linewidth}
\begin{tabular}{@{}lccccc@{}}
\toprule
\textbf{Feature} & \textbf{GCN} & \textbf{SAGE} & \textbf{GIN} & \textbf{GINE} & \textbf{D-GCN} \\
\midrule
Normalized aggregation     & \cmark & \cmark & \xmark & \xmark & \xmark \\
Per-neighbor weighting     & \xmark & \xmark & \xmark & \cmark & \cmark \\
Edge-aware messages        & \xmark & \xmark & \xmark & \cmark & \xmark \\
Nonlinearity \emph{before} aggregation & \xmark & \xmark & \xmark & \cmark & \cmark \\
Per-layer MLPs             & \xmark & \xmark & \cmark & \cmark & \xmark \\
Decoupled self vs.\ neighbor paths  & \xmark & \cmark & \xmark & \xmark & \cmark \\
Heterophily robustness     & Low    & Medium & Medium & High   & High  \\
Over-smoothing resistance  & Low    & Medium & Medium    & Med–High & High \\
\bottomrule
\end{tabular}
\end{adjustbox}
\end{table}

D-GCN achieves 3.3\% NMAE compared to 63.94\% for standard GCN, demonstrating that domain-specific 
architectural design, decoupled self/neighbor processing with learnable attention and unnormalized 
aggregation—is essential for capturing wireless interference dynamics that generic GNN architectures 
cannot model effectively.

\section{Dataset Generation and Evaluation Metrics}
To train, validate, and benchmark the proposed GNN, we require
\emph{per–topology, per–node} ground-truth throughput labels. 

We generate data by running a p-CSMA network under two models:

\subsubsection*{1) Event‐Driven Simulation (Approximate)}
We first create random Erd\H{o}s–R\'enyi conflict graphs of size
\(n \in \{3,4,\dots,20\}\) with edge-creation probability
\(p_{\text{edge}} = 0.5\) (i.e., a fresh random topology for each run). Erd\H{o}s–R\'enyi intentionally creates 
diverse conflict graph structures to prevent overfitting to specific spatial layouts. Since 
D-GCN operates on conflict graph topology rather than physical coordinates \cite{lu2024graph, 
li2017conflict}
Every node is assigned an independent access probability
\(p_i \sim \mathcal{U}(0,1)\).
On this topology, we run a saturated p-CSMA,
event–driven timeline of \(10^{6}\) time slots:
a node that wins the channel occupies it for \(T\) consecutive
slots, while any node that collides waits \(\sigma\) idle slots
(\(\sigma = 1\) in all experiments) before re-contending.
By counting collision-free transmission starts we compute each node’s
throughput as
\(\Theta_i^{\text{Sim}}
      = \bigl(\#\text{succ}_i \times T\bigr) / 10^{6}\),
giving the approximate throughput vector
\(\boldsymbol{\Theta}^{\text{Sim}}\).
\paragraph*{Simulation precision}
Each simulation run spans $L=10^{6}$ time slots to ensure statistical reliability. The per-node throughput estimate $\hat{\Theta}_i = (S_i T)/L$, where $S_i$ counts collision-free transmission starts, exhibits Monte Carlo sampling error. Under Poisson approximation for large $L$, the standard error is $\mathrm{SE}(\hat{\Theta}_i) \approx \sqrt{\hat{\Theta}_i T/L}$, yielding 95\% confidence intervals of $\hat{\Theta}_i \pm 1.96\,\mathrm{SE}(\hat{\Theta}_i)$.

While the resulting confidence half-widths are small in absolute terms---$2.77 \times 10^{-3}$ for $T=2$ and $5.54 \times 10^{-3}$ for $T=8$---they become relevant when working with throughput values on similar scales. For instance, with typical throughputs around $0.01$--$0.1$, these uncertainties represent 3--28\% relative error for $T=2$ and 6--55\% for $T=8$. This simulation-induced variance contributes measurably to the model's prediction error, particularly for longer transmission durations and number of nodes, where both throughput values and uncertainties are smaller.

\subsubsection*{2) Markov-Chain Solver (Exact)}
The same network can be analysed exactly by modelling it as a global
discrete-time Markov chain whose state at slot~\(t\) is the vector of
\emph{remaining busy times}
\(\mathbf{a}(t) = (a_1,\dots,a_n) \in \{0,\dots,T-1\}^{n}\).
There are \(T^{\,n}\) such states; for each state we enumerate all
\(2^{\,n}\) feasible transmission decisions, build the transition
matrix \(P\), and solve
\(P^{\!\top}\boldsymbol{\pi} = \boldsymbol{\pi}\)
for the stationary distribution \(\boldsymbol{\pi}\).
Using the reward decomposition in~\cite{11133870} we obtain the exact
per-node saturation throughput vector
\(\boldsymbol{\Theta}^{\text{MC}}\).

We repeat the above procedure for many independently generated random topologies; each iteration run writes one row to a CSV file containing the graph’s adjacency matrix, the per-node access-probability vector, and the resulting saturation-throughput vector, providing a reusable dataset for subsequent analysis.

\subsection{Graph-Neural Network Configuration}
For each network topology, we construct an undirected conflict graph $G=(V,E)$, 
where vertices represent wireless transmitters and edges encode pairwise interference relationships. 
Each node $i \in V$ is initialized with a feature vector $\mathbf{x}_i = [p_i]$ containing its transmission probability. 
We also experimented with augmented features $\mathbf{x}_i = [p_i, T]$ that include transmission duration, 
though these yielded only marginal improvements.

As illustrated in Figure~\ref{fig:Architecture}, our GNN architecture processes the input graph 
through multiple stacked D-GCN layers, where each layer aggregates information from immediate neighbors. 
With $L$ layers in the D-GCN stack, the model can theoretically capture interference dependencies up to $L$ hops away—an 
important property for modeling cascading effects.

For implementation, we use:
- \textbf{Architecture}: 8 D-GCN layers (7 hidden layers + 1 additional layer) with 64 hidden units each, followed by a 2-layer MLP head [64 → 32 → 1]
- \textbf{Activation}: ReLU for hidden layers, sigmoid for final output
- \textbf{Training}: AdamW optimizer with learning rate 0.001 and weight decay $10^{-4}$
- \textbf{Learning rate scheduling}: ReduceLROnPlateau with factor 0.5 and patience 5
- \textbf{Loss function}: MSE for training, with MAE and NMAE for evaluation
- \textbf{Gradient clipping}: Maximum norm of 1.0 to ensure stable training

This architecture effectively balances model expressiveness with computational efficiency, making it suitable for real-world wireless network optimization tasks.

Our D-GCN models consistently converge within 150-200 epochs across all experimental configurations. This rapid convergence is typical for GNN architectures on moderately-sized graphs, as the local message passing mechanism efficiently propagates information through the network structure \cite{dwivedi2023benchmarking}.

\subsection{Evaluation Metrics}

We assess model fidelity by comparing the predicted throughputs $\hat{\Theta}_i$ directly against the ground-truth saturation throughputs \(\Theta_i\).

\begin{enumerate}[label=\alph*)]
\item \textbf{Mean Squared Error (MSE)}  
      $\text{MSE}
      =\tfrac1{N}\sum_{i}(\hat{\Theta}_i-\Theta_i)^{2}$  
      is used as the \emph{training loss} because it
      provides smooth gradients and heavily penalises large mistakes.

\item \textbf{Mean Absolute Error (MAE)}  
      $\text{MAE}
      =\tfrac1{N}\sum_{i}\lvert\hat{\Theta}_i-\Theta_i\rvert$  
      offers an interpretable ``average mistake’’ in throughput units.

\item \textbf{Normalised MAE (NMAE)}  
      $\text{NMAE}= \text{MAE}\big/\overline{\Theta}$,  
      where $\overline{\Theta}$ is the sample mean of ground-truth
      throughput, reports the \emph{relative} error and enables fair
      comparison across datasets with different settings.
\end{enumerate}

\section{Experimental Results \& Performance Evaluation}
This section presents a comprehensive evaluation of the proposed D-GCN model, focusing on its predictive accuracy, generalization ability, and computational efficiency. 
We compare D-GCN against multiple GNN baselines, analyze its robustness to different network configurations, and assess its effectiveness in gradient-based utility optimization.

\subsection{Performance comparison with other GNN architectures}
To evaluate the effectiveness of our proposed D-GCN architecture, we conducted a comprehensive comparison against several state-of-the-art GNN models on the throughput prediction task. All architectures were trained on the same dataset with packet duration $T=5$ and evaluated under identical test configurations to ensure a fair comparison.  

Table~\ref{tab:mae_nmae} summarizes the test-set performance of five GNN variants: Graph Convolutional Network (GCN), GraphSAGE, Graph Isomorphism Network (GIN), Graph Isomorphism Network with Edge Features (GINE), and the proposed Decoupled Graph Convolutional Network (D-GCN). Each model uses only the transmission probability $p_i$ as the node feature, isolating the impact of architectural differences on learning the nonlinear mapping from local transmission probabilities to global throughput outcomes.  

Our D-GCN achieves the lowest normalized mean absolute error (NMAE) of 3.3\%, significantly outperforming GCN (63.9\%), GraphSAGE (23.7\%), GIN (21.4\%), and GINE (4.7\%). These results highlight that D-GCN’s decoupled self/neighbor design and unnormalized attention aggregation enable it to capture the nonlinear interference relationships in wireless networks far more effectively than standard GNN architectures.

\begin{table}[h]
\centering
\caption{Test-set error of evaluated GNN architectures
         ($T{=}5$ dataset, single node feature $p_i$).}
\label{tab:mae_nmae}
\begin{tabular}{lcc}
\toprule
\textbf{Architecture} & \textbf{MAE} & \textbf{NMAE} \\
\midrule
GCN                & 0.0495 & 0.6394 \\
SAGE        & 0.0183 & 0.2372 \\
GIN                & 0.0165 & 0.2135 \\
GINE               & 0.0037 & 0.0470 \\
\textbf{D-GCN (ours)} & \textbf{0.0026} & \textbf{0.0330} \\
\bottomrule
\end{tabular}
\end{table}

\begin{figure}[h!]
    \centering
    \includegraphics[width=0.7\linewidth]{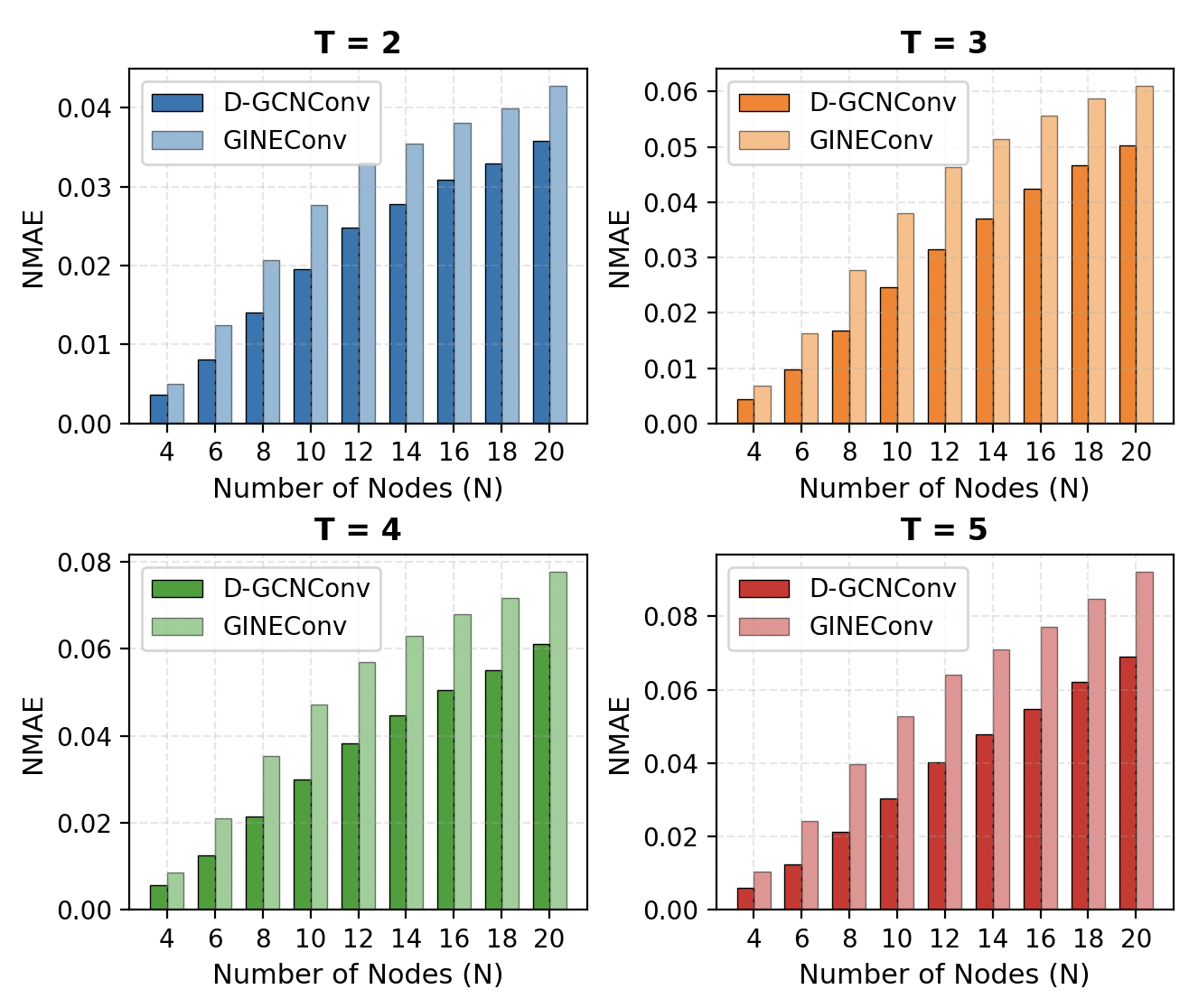}
    \caption{Comparison of Normalized Mean Absolute Error (NMAE \%) between D-GCN (Decoupled GCN) and GINE architectures across different transmission durations (T = 2, 3, 4, 5) and network sizes ($N\in\{4,6,8,10,12,14,16,18,20\}$)}
    \label{figDgcnGine}
\end{figure}

Figure~\ref{figDgcnGine} demonstrates the consistent superiority of our proposed D-GCN architecture over GINE across all tested configurations, with simpler, more interpretable operations that align with wireless physics.

\subsection{Performance with Different Number of Training Samples}
To evaluate the data efficiency of our D-GCN architecture, we conducted experiments varying the training dataset size. We generated 5,000 graphs for each network size ($N\in\{4,6,8,10,12,14,16,18,20\}$) at $T=5$, creating a total dataset of 45,000 samples. We then trained models using 10\%, 25\%, 50\%, 75\%, and 100\% of this dataset. Table~\ref{tab:training_samples} shows the test performance for each training set size. The model achieves reasonable performance (NMAE $<=$ 8\%) with as few as 4,500 samples (10\% of data), demonstrating efficient learning of the underlying throughput dynamics. Performance improves substantially from 10\% to 50\% of the data, with NMAE decreasing from 7.30\% to 4.06\%. Beyond 50\%, the gains become marginal—using the full dataset only reduces NMAE by an additional 0.74\%. This rapid convergence with limited data is particularly valuable for practical deployments where generating ground-truth labels through Markov chain analysis or extensive simulations is computationally expensive. The consistent gap between training and test NMAE across all dataset sizes indicates good generalization without overfitting.

\begin{table}[h]
\centering
\caption{Model performance with varying training dataset sizes ($T=5$, $N\in\{4,6,\dots,20\}$)}
\label{tab:training_samples}
\begin{tabular}{lccc}
\toprule
\textbf{Training Data} & \textbf{Train NMAE (\%)} & \textbf{Test NMAE (\%)} \\
\midrule
10\% (4500)    & 6.81 & 7.30 \\
25\%    & 3.99 & 4.81 \\
50\% (22,500)    & 3.37 & 4.06 \\
75\%    & 2.63 & 3.24 \\
100\% (45,000)   & 2.86 & 3.32 \\
\bottomrule
\end{tabular}
\end{table}

\subsection{Performance with Different Network Settings}
\begin{figure}[h!]
    \centering
    \includegraphics[width=0.55\linewidth]{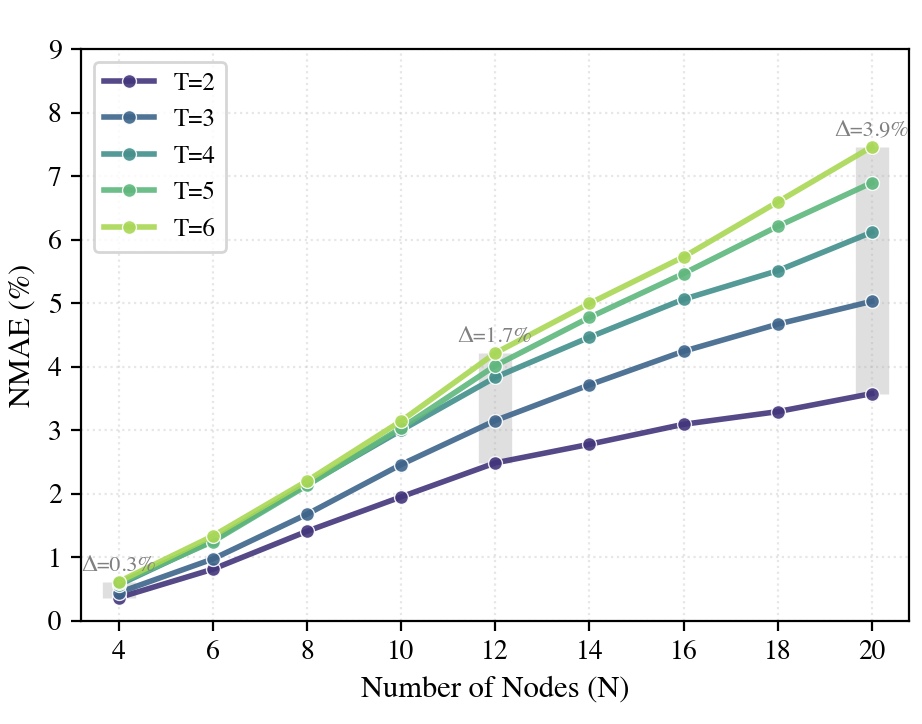}
  \caption{Performance of the D-GCN across different network configurations, varying in the number of nodes and transmission durations.}
    \label{fig:diffnetworksetting}
\end{figure}

Figure~\ref{fig:diffnetworksetting} illustrates the D-GCN model's performance across diverse network configurations, varying both transmission duration ($T\in\{2,3,4,5,6\}$) and network size ($N\in\{4,6,8,10,12,14,16,18,20\}$). The results reveal two key trends. First, prediction accuracy decreases as network size increases, with NMAE rising from 0.36\%--0.61\% for 4-node networks to 3.58\%--7.46\% for 20-node networks. This degradation has two causes: larger graphs exhibit more complex multihop interference patterns that are inherently harder to model, and larger networks require longer simulation times to reach steady state, though we fixed all simulations at one million time slots.

Second, model performance was assessed against simulation uncertainty bounds. For $T=2$, the test NMAE was $1.64\% \pm 0.11\%$ (95\% CI), where the confidence interval reflects propagated simulation uncertainty. The maximum simulation-induced relative uncertainty was 0.28\%, substantially smaller than the 1.64\% model error. Similarly, for $T=8$, the test NMAE of $4.30\% \pm 0.24\%$ greatly exceeded the 0.55\% maximum simulation uncertainty. These results demonstrate that model errors are dominated by approximation rather than simulation noise, with error-to-uncertainty ratios of approximately 6:1 and 8:1 for $T=2$ and $T=8$, respectively.

Our experiments on networks up to 20 nodes 
provide comprehensive validation of D-GCN's ability to capture local interference patterns, 
which is the fundamental challenge in heterogeneous p-CSMA. The architecture itself is not 
constrained by network size, it processes graphs through local message passing with complexity 
$O(|E| \cdot d \cdot L)$, enabling efficient inference on larger networks through inductive 
generalization \cite{hamilton2017inductive}. Real wireless deployments exhibit localized 
interference neighborhoods of 10-15 nodes despite containing hundreds of devices 
\cite{abdallah2024improving, cilfone2019wireless}, meaning our experimental scale captures 
the relevant dynamics.
Notably, the performance curves converge for larger $T$ values ($T \geq 5$)—the NMAE difference 
between $T=5$ and $T=6$ is minimal compared to $T=2$ versus $T=3$. This reflects the underlying 
p-CSMA dynamics, as transmission duration increases, longer channel occupancy periods dominate 
the throughput calculation, making the relative impact of $T$ less significant. The model 
accurately captures this inherent property of the protocol.

\subsection{Generalizability to Different Network Settings}
The generalization capability of our D-GCN model was evaluated across two dimensions: network size and transmission duration. Table~\ref{tab:gen_holdoutN} shows the model's performance when trained on small graphs ($N\in\{4,6,8,10,12\}$) and tested on larger networks ($N\in\{14,16,18,20\}$). The NMAE increases progressively from 6.90\% to 15.38\% at $N=20$, reflecting the inherent difficulty of extrapolating to networks with more complex multihop interference patterns. Notably, when trained on the full range of network sizes ($N\in[4,20]$), the NMAE remains below 7\% across all test cases, demonstrating that comprehensive training data significantly improves generalization. Table~\ref{tab:gen_holdoutT} reveals strong temporal generalization, the model achieves NMAE of 4.58\% and 6.29\% on unseen transmission durations $T\in\{7,8\}$ when trained only on $T\in\{2,3,4,5,6\}$. This asymmetric generalization, stronger for temporal parameters than spatial configurations, aligns with the fundamental nature of p-CSMA networks, where temporal dynamics follow predictable protocol behavior while spatial interference patterns grow exponentially with network size. These results confirm that our D-GCN architecture effectively captures the underlying throughput dynamics, making it suitable for practical deployment in heterogeneous wireless networks.

\begin{table}[h!]
\centering
\caption{Generalization across network size at $T{=}5$. Left column: model trained on small graphs ($N\in\{4,6,8,10,12\}$) and tested on larger graphs ($N\in\{14,16,18,20\}$). Right column: model trained on all sizes ($N\in[4,8,10,12,14,16,18,20]$) and tested on ($N\in\{14,16,18,20\}$). We report normalized MAE (NMAE).}
\label{tab:gen_holdoutN}
\setlength{\tabcolsep}{6pt}
\small
\begin{tabular}{@{}rcc@{}}
\toprule
 & \textbf{Train $N=4,6$--$12$} & \textbf{Train $N=4,6$--$20$} \\
\cmidrule(l){2-3}
\textbf{$N$} & \textbf{NMAE (\%)} & \textbf{NMAE (\%)} \\
\midrule
14  & 6.25  & 4.77 \\
16  & 8.59  & 5.47 \\
18  & 12.09 & 6.21 \\
20  & 15.38 & 6.90 \\
\bottomrule
\end{tabular}
\end{table}

\begin{table}[h]
\centering
\caption{Generalization across transmission duration $T$. Left column: model trained on $T\in\{2,3,4,5,6\}$ and tested on unseen $T\in\{7,8\}$. Right column: model trained on $T\in\{2,\dots,8\}$. (All models were trained on mixed sizes $N\in[4,20]$.) We report normalized MAE (NMAE).}
\label{tab:gen_holdoutT}
\setlength{\tabcolsep}{6pt}
\small
\begin{tabular}{@{}rcc@{}}
\toprule
 & \textbf{Train $T{=}2$--$6$} & \textbf{Train $T{=}2$--$8$ (all-$T$)} \\
\cmidrule(l){2-3}
\textbf{$T$} & \textbf{NMAE (\%)} & \textbf{NMAE (\%)} \\
\midrule
7 & 4.58 & 3.31 \\
8 & 6.29 & 3.59 \\
\bottomrule
\end{tabular}
\end{table}

\subsection{Computational Efficiency Analysis}
To quantify D-GCN's computational advantage over exact Markov chain analysis, we conducted systematic timing experiments across network configurations of varying complexity. All experiments were performed on a MacBook Pro with an Apple M2 Pro processor.

Table~\ref{tab:computational_efficiency} demonstrates D-GCN's decisive computational efficiency. While the exact Markov chain method exhibits exponential scaling with state space size $O(T^n \cdot 2^n)$, D-GCN maintains near-constant inference time across all configurations. The Markov analysis becomes computationally intractable for networks with 10 nodes at $T=3$ (requiring enumeration of 59,049 states), whereas D-GCN completes inference in under one millisecond. This efficiency translates to speedups ranging from 3× for small networks to over 195,000× for larger configurations, enabling real-time optimization applications that would be infeasible with exact methods.

\begin{table}[h]
\centering
\caption{Computation time comparison between exact Markov chain analysis and D-GCN inference across different network configurations.}
\label{tab:computational_efficiency}
\resizebox{0.7\columnwidth}{!}{%
\begin{tabular}{ccrrrr}
\toprule
\textbf{Nodes} & \textbf{T} & \textbf{State Space} & \textbf{MC Time (s)} & \textbf{D-GCN Time (s)} & \textbf{Speedup} \\
\midrule
\rowcolor{gray!10}
5  & 2 & 32      & 1.82 × 10$^{-3}$   & 6.30 × 10$^{-4}$ & 2.9× \\
5  & 3 & 243     & 2.96 × 10$^{-2}$   & 6.47 × 10$^{-4}$ & 45.7× \\
\rowcolor{gray!10}
6  & 2 & 64      & 1.44 × 10$^{-2}$   & 6.36 × 10$^{-4}$ & 22.7× \\
6  & 3 & 729     & 1.43 × 10$^{-1}$   & 6.62 × 10$^{-4}$ & 216.8× \\
\rowcolor{gray!10}
7  & 2 & 128     & 4.58 × 10$^{-2}$   & 6.89 × 10$^{-4}$ & 66.5× \\
7  & 3 & 2,187   & 1.01              & 7.13 × 10$^{-4}$ & 1,412× \\
\rowcolor{gray!10}
8  & 2 & 256     & 1.76 × 10$^{-1}$   & 7.21 × 10$^{-4}$ & 244.8× \\
8  & 3 & 6,561   & 8.72              & 7.30 × 10$^{-4}$ & 11,943× \\
\rowcolor{gray!10}
9  & 2 & 512     & 7.29 × 10$^{-1}$   & 7.34 × 10$^{-4}$ & 993.5× \\
9  & 3 & 19,683  & 145.31            & 7.42 × 10$^{-4}$ & 195,770× \\
\rowcolor{gray!10}
10 & 2 & 1,024   & 2.55              & 7.53 × 10$^{-4}$ & 3,387× \\
10 & 3 & 59,049  & \textit{intractable}$^{\dagger}$ & 7.80 × 10$^{-4}$ & -- \\
\bottomrule
\multicolumn{6}{l}{\footnotesize $^{\dagger}$Process terminated after exceeding 1000s runtime threshold.}
\end{tabular}
}
\end{table}

The results reveal two critical insights. First, D-GCN's inference time remains remarkably stable (6.3–7.8 × 10$^{-4}$ seconds) regardless of network size or packet duration, demonstrating $O(1)$ complexity with respect to the state space. Second, the speedup factor increases exponentially with network complexity, making D-GCN particularly valuable for large-scale network optimization where hundreds of throughput evaluations are required.

\subsection{Application to Network Utility Maximization}
To further evaluate the practical utility of the proposed D-GCN model, we examine its ability to support gradient-based optimization of network parameters. In this experiment, D-GCN is embedded within an end-to-end utility maximization loop, where node transmission probabilities are iteratively adjusted using stochastic gradient descent (SGD) to maximize a weighted network utility function. Our objective is to determine optimal transmission probabilities $\mathbf{p}$ that maximize the utility function:
\[
J(\mathbf{p})=\sum_i \alpha_i \log(\Theta_i(\mathbf{p})+\varepsilon), \quad \varepsilon=10^{-9},
\]
where $\alpha_i$ represents the utility weight for node $i$ and $\Theta_i(\mathbf{p})$ denotes its throughput.

We compare two optimization approaches:
\begin{enumerate}
    \item \textbf{Markov (Exact):} Projected gradient ascent on $J$ using central finite-difference gradients computed from the exact Markov chain model.
    \item \textbf{D-GCN (Learned):} Gradient ascent utilizing the pre-trained D-GCN to predict throughput $\Theta_i(\mathbf{p})$, with gradients obtained via backpropagation.
\end{enumerate}

Both methods employ identical initialization $\mathbf{p}_{\text{init}}$, learning rates, and probability constraints within $[0,1]$.

\begin{figure}[h!]
    \centering
    \includegraphics[width=0.55\linewidth]{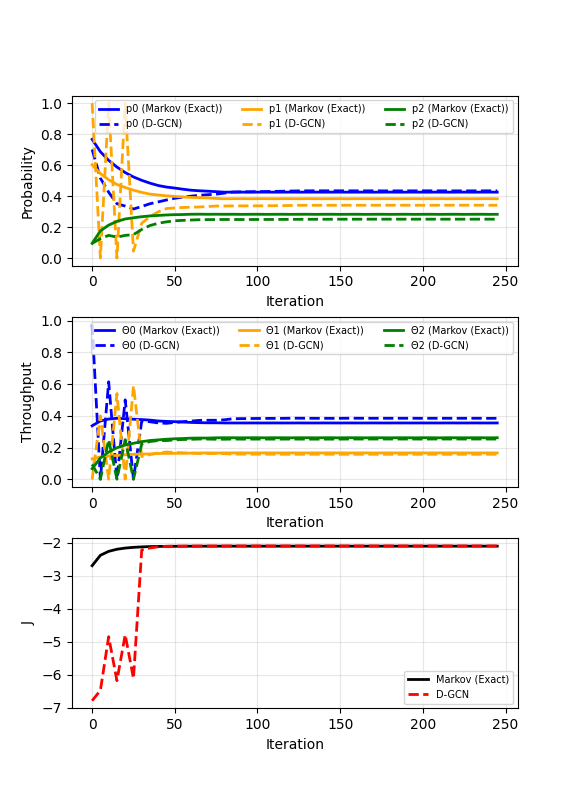}
    \caption{Optimization trajectories comparing Markov model (solid) and D-GCN (dashed). Top panel: transmission probabilities; middle panel: per-node throughput; bottom panel: utility function $J$.}
    \label{fig:utility_comparison}
\end{figure}

Figure~\ref{fig:utility_comparison} demonstrates that D-GCN closely replicates the exact model's optimization trajectory on a 3-node chain topology ($0 \leftrightarrow 1 \leftrightarrow 2$). With initial probabilities $\mathbf{p}_{\text{init}}=[0.97,\,0.01,\,0.05]$, utility weights $\boldsymbol{\alpha}=[0.6,\,0.6,\,0.3]$, and SGD optimization (learning rate $0.01$), the final utilities differ by less than 1\%. These results confirm that D-GCN not only predicts throughput accurately but also enables efficient, differentiable optimization of network parameters consistent with analytical solutions.


\begin{figure}[h!]
    \centering
    \includegraphics[width=0.9\linewidth]{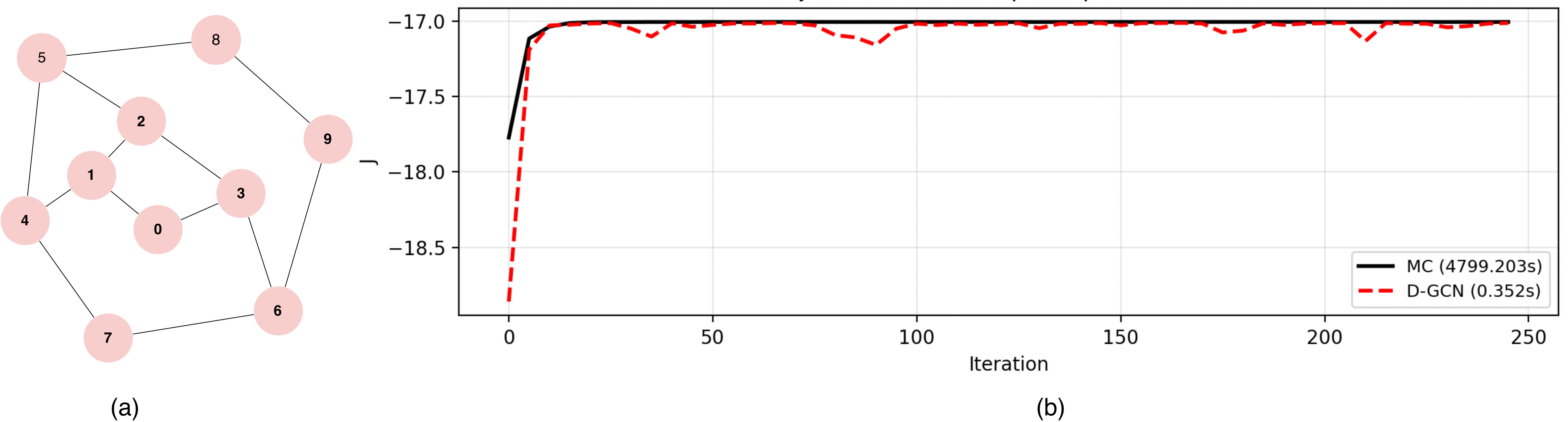}
        \caption{(a) Conflict graph of the 10-node CSMA network. (b) Convergence of D-GCN (red dashed) and Markov chain baseline (black solid) on the same network, showing similar steady-state utility $J$.}
    \label{fig:10_optimization}
\end{figure}

To evaluate the scalability of our approach, we tested both methods on a 10-node network with a more complex interference structure (Figure~\ref{fig:10_optimization}(a)), representing a realistic wireless deployment scenario.
The optimization uses initial probabilities $\mathbf{p}_{\text{init}} \in [0.10, 0.30]$ with heterogeneous utility weights $\boldsymbol{\alpha} \in [0.8, 1.1]$ to reflect diverse QoS requirements. Both methods employ SGD with learning rate 0.01 over 250 iterations, optimizing the same log-utility objective with packet duration $T=2$.

Table~\ref{tab:num10} presents the optimization results of both methods to achieve virtually identical final utilities (difference $< 0.05\%$). The variation in equilibrium probabilities between methods reflects the presence of multiple local optima, a characteristic feature of CSMA networks. Importantly, D-GCN's ability to identify an equivalent-quality solution validates its effectiveness as an efficient alternative for large scale network optimization problems. As shown in Figure~\ref{fig:10_optimization}(b), both approaches achieve nearly identical utility values (D-GCN: $J=-17.013$, MC: $J=-17.006$) after 250 iterations, validating that our D-GCN accurately captures the p-CSMA dynamics even in larger networks.

\begin{table}[h!]
\centering
\caption{Optimization results for the 10-node network after 250 iterations.}
\label{tab:num10}
\resizebox{0.5\columnwidth}{!}{%
\begin{tabular}{ccccccc}
\toprule
\multirow{2}{*}{Node} & \multicolumn{2}{c}{MC Optimization} & \multicolumn{3}{c}{D-GCN Optimization} \\
\cmidrule(lr){2-3} \cmidrule(lr){4-6}
& $p_i^{\text{MC}}$ & $\Theta_i^{\text{MC}}$ & $p_i^{\text{D-GCN}}$ & $\Theta_i^{\text{D-GCN}}$ & $\Theta_i^{\text{MC-eval}}$ \\
\midrule
0 & 0.2697 & 0.1919 & 0.3189 & 0.2533 & 0.2539 \\
1 & 0.2911 & 0.1624 & 0.2677 & 0.1239 & 0.1137 \\
2 & 0.2525 & 0.1300 & 0.3079 & 0.1778 & 0.1868 \\
3 & 0.3116 & 0.1795 & 0.2855 & 0.1280 & 0.1216 \\
4 & 0.2782 & 0.1319 & 0.3007 & 0.1736 & 0.1788 \\
5 & 0.2801 & 0.1486 & 0.2689 & 0.1224 & 0.1122 \\
6 & 0.2523 & 0.1006 & 0.3146 & 0.1644 & 0.1631 \\
7 & 0.3038 & 0.2487 & 0.2431 & 0.1809 & 0.1739 \\
8 & 0.3082 & 0.1976 & 0.3355 & 0.2475 & 0.2470 \\
9 & 0.3360 & 0.2432 & 0.2837 & 0.1840 & 0.1768 \\
\midrule
\textbf{Final $J$} & \multicolumn{2}{c}{-17.0057} & \multicolumn{2}{c}{-17.0131} & -17.0285$^*$ \\
\bottomrule
\multicolumn{6}{l}{\footnotesize $^*$Utility calculated using MC model with D-GCN optimized probabilities.}
\end{tabular}
}
\end{table}

 The most striking advantage of D-GCN lies in its computational efficiency. While both methods converge to comparable solutions, the time required differs by four orders of magnitude, the Markov Chain method requires 4,799.2 seconds (approximately 80 minutes) to complete the optimization, whereas D-GCN achieves the same result in just 0.352 seconds, a remarkable 13,621× speedup. This dramatic speedup stems from the fundamental difference in computational approach, the Markov chain method requires solving a system with $\mathcal{O}(T^n)$ states and computing stationary distributions at each gradient step, while D-GCN performs a single forward pass through the trained network with $\mathcal{O}(n)$ complexity. This computational advantage makes D-GCN practical for real-time optimization in dynamic wireless networks, where rapid adaptation to changing conditions is crucial.

\section{Code and Data Availability}
The source code for the D-GCN model, dataset generation scripts, and experimental configurations are publicly available at \url{https://github.com/ANRGUSC/predictCSMA}. The repository includes implementation details, hyperparameter settings, and instructions for reproducing the experimental results.

\section{Conclusions}

This paper presents the first Graph Neural Network application for predicting per-node saturation throughput in heterogeneous p-CSMA networks, addressing the computational intractability of exact Markov methods. Our Decoupled Graph Convolutional Network (D-GCN) introduces an interpretable architecture that separates self-transmission from neighbor interference without degree normalization, using learnable attention weights to capture heterogeneous neighbor impacts. D-GCN achieves 3.3\% NMAE versus 63.94\% for standard GCN while maintaining interpretability about interference sources.


By providing differentiable throughput estimates, D-GCN enables gradient-based network optimization that achieves utility within 1\% of theoretical optima while offering computational speedups of 100-1000× compared to exact Markov chain methods.

Several limitations merit acknowledgment. First, while D-GCN handles networks up to 20 nodes effectively, scalability to larger networks (50+ nodes) remains unexplored. Second, the model assumes saturated traffic conditions, where all nodes continuously have packets to transmit. Real-world networks often exhibit non-saturated, time-varying traffic patterns with bursty arrivals and idle periods. Extending our approach to these scenarios would require incorporating queue state information and temporal dynamics into the node features, along with generating appropriate training data that captures diverse traffic conditions. Finally, our approach requires ground-truth labels from either expensive simulations or exact analytical methods for training, though we demonstrated that as few as 4,500 samples suffice for reasonable performance.

Future research directions include: (i) extending the architecture to handle non-saturated traffic patterns and variable packet lengths, (ii) incorporating physical layer parameters such as signal-to-interference ratios and channel conditions, (iii) developing online learning mechanisms that adapt to dynamic network conditions, (iv) investigating the application of our decoupled architecture to other MAC protocols beyond p-CSMA and other use cases.

In conclusion, this work demonstrates that carefully designed GNN architectures can serve as accurate, efficient surrogate models for complex wireless protocol analysis.

\bibliographystyle{ACM-Reference-Format}

\bibliography{sample-base}

\end{document}